\documentclass[runningheads]{llncs}
\usepackage[T1]{fontenc}
\usepackage{graphicx}
\usepackage{natbib} 
\usepackage{multirow}
\usepackage{hyperref}
\begin{document}
\title{Word Sense Disambiguation as a Game of Neurosymbolic Darts}
%
%
\author{Tiansi Dong  \and
Rafet Sifa}
\authorrunning{T.Dong and R.Sifa}
%
\institute{Fraunhofer Institute IAIS  \\
\email{\{tiansi.dong,rafet.sifa\}@iais.fraunhofer.de}}
\maketitle              
\begin{abstract} 
Word Sense Disambiguation (WSD) is one of the hardest tasks in natural language understanding and knowledge engineering. The glass ceiling of 80\% F1 score is recently achieved through supervised deep-learning, enriched by a variety of knowledge graphs. Here, we propose a novel neurosymbolic methodology that is able to push the F1 score above 90\%. The core of our methodology is a neurosymbolic sense embedding, in terms of a configuration of nested balls in $n$-dimensional space. The centre point of a ball well-preserves word embedding, which partially fix the locations of balls. Inclusion relations among balls precisely encode symbolic hypernym relations among senses, and enable simple logic deduction among sense embeddings, which cannot be realised before. We trained a Transformer to learn the mapping from a contextualized word embedding to its sense ball embedding, just like playing the game of darts (a game of shooting darts into a dartboard). A series of experiments are conducted by utilizing pre-training $n$-ball embeddings, which have the coverage of around $70\%$ training data and $75\%$ testing data in the benchmark WSD corpus. The F1 scores in experiments range from $90.1\%$ to $100.0\%$ in all six groups of test data-sets (each group has 4 testing data with different sizes of $n$-ball embeddings). Our novel neurosymbolic methodology has the potential to break the ceiling of deep-learning approaches for WSD. Limitations and extensions of our current works are listed. 
\keywords{word sense disambiguation  \and neurosymbolic representation \and knowledge graph \and natural language understanding.}
\end{abstract}

\section{Introduction}

Word Sense Disambiguation (WSD) is the task of acquiring the intended meaning of
a word within the context where it appears \citep{Navigli09}. WSD is one of the fundamental topics of natural language understanding in Artificial Intelligence (AI), which can date back to 1949 \citep{weaver49}, in part because WSD affects a variety of downstream AI applications, such as information extraction, entailment analysis, machine translation, opinion mining, question-answering,  sentiment analysis, text understanding. By hybridizing knowledge graphs and pre-trained language models, deep neural-network approaches have attained estimated human performance, and reached a glass ceiling over $80\%$ \citep{navigli21survey}, yet, they still make simple mistakes that
humans would not do \citep{navigli2022}. 
 \begin{figure}[t]
\centering  
\includegraphics[width=1\textwidth]{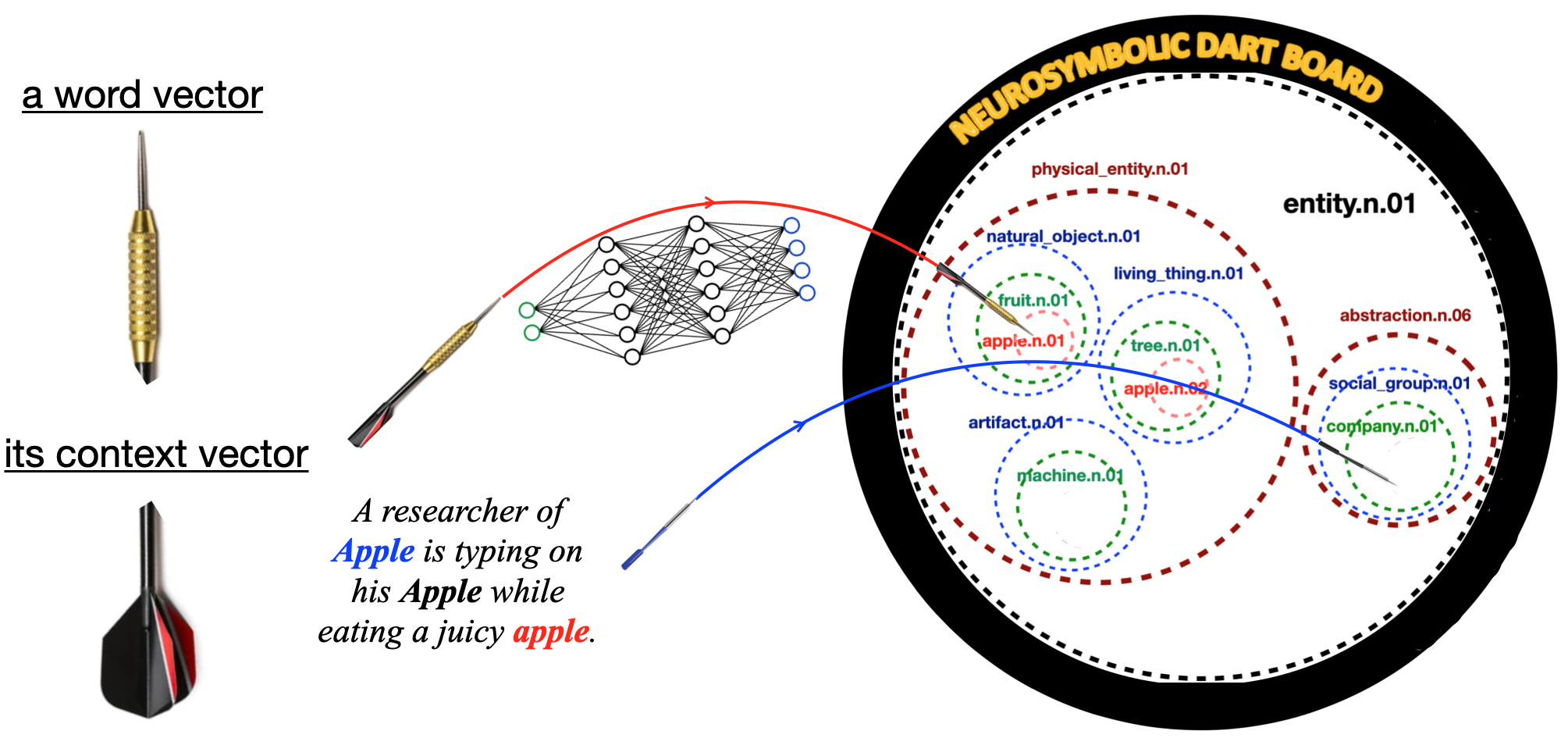} \caption{A neurosymbolic approach to Word-Sense Disambiguation works like playing the game of Dart. A deep neural network learns to shoot a contextualized word embedding vector to its sense regions in the Dart board. 
}
\label{fig:frame}
\end{figure}
Neural approaches assume that a word-sense is an {\em opaque} class, and that  classifying a word and its context into a word-sense class is limited to the knowledge that can be acquired from the training data \citep{navigli21survey}. This hypothesis is not only theoretically sound, but also practical, within the traditional deep learning paradigm. However, recent progress in neurosymbolic research is able to {\em explicitly} represent such {\em opague} sense class in terms of a region configuration. For example, a large symbolic tree structured taxonomy of word-senses can be precisely merged with word embeddings without any loss \citep{dong19iclr,dong19,DongSpringer21}. That is, in the neurosymbolic paradigm, a word-sense is no more an {\em opaque} class, rather that it can be explicitly embedded as an $n$-dimensional ball with a crisp boundary. This provides a new way to tackle the tough WSD problem, which can be vividly imagined as a game of dart as follows: A neurosymbolic WSD is a neural dart player that shoots a contextualized word vector to the place of a configuration of regions, where its sense-region is located. This configuration of regions precisely encodes the sense inventory and latent features of words, as illustrated in Figure~\ref{fig:frame}.

For example, apple.n.01, orange.n.01, and watermelon.n.01 are members of fruit.n.01. In classic deep-learning approaches, they are embedded as four vectors that lack the representation power to explicitly capture such membership relations. But, if we represent them as regions, for example, balls, things will change, the membership relations will be explicitly represented by inclusion relations among balls: the balls of apple.n.01, orange.n.01, and watermelon.n.01 are inside the ball of fruit.n.01. The advantage for WSD is not only that shooting to a region  is much easier than shooting to a point, but also that explicit ball representation enables logical deduction among senses: 
shooting a contextualized vector of the word {\em apple} to the ball of
fruit.n.01 is sufficient to determine apple.n.01 as the intended sense; while
shooting to the ball of abstraction.n.06 is reasonable to hypothesize that it may refer to a company, even although {\em apple} does not have the sense of company ({\em company.n.01}) in the sense inventory, as shown by the blue shooting path in Figure~\ref{fig:frame}. 

Our contribution in this paper are listed as follows.
\begin{enumerate}
    \item We propose a novel neurosymbolic methodology for WSD, which seamlessly unifies supervised learning approaches and simple symbolic reasoning among hypernym relations. 
    \item We implement the first neurosymbolic WSD system, by assemblizing pre-trained word-embeddings and pre-trained $n$-ball embeddings of word-senses as the input and the output of a Transformer. Its performance breaks the ceiling of traditional deep-learning approached in the current experiment setting.
    \item Our experiment results provide valuable lessons for continued research for full-fledged neurosymbolic multi-lingual WSD systems. All source codes and data-sets are open, in order to promote research and collaborations. 
\end{enumerate} 

The rest of the article is structured as follows: Section 2 reviews the recent WSD methods, and shows relations to our approach; Section 3 introduces our novel neurosymbolic approach in detail; Section 4 reports the experiments and results, which greatly outperforms the best neural approach for WSD; Section 5 concludes the article and discusses a number of future works. 

\section{Related Works}

\subsection{Word Sense Disambiguation}


The task of Word Sense Disambiguation (WSD) is to automatically decide the intended sense in a given context, where senses of words are selected from the fixed word-sense inventory \citep{navigli21survey}.
A WSD system has three components, as follows: (1) a word in a given context, (2) a word sense inventory, e.g., WordNet \citep{Miller90,MillerWordNet95}, BabelNet \citep{Navigli21BabelNet}, and (3) an annotated corpus, e.g. SemCor \citep{semcor93}, where some words have been manually or automatically annotated with intended word senses.
The knowledge graph approaches and  supervised deep-learning approaches  are the main WSD approaches. Their performances are determined by the quality and the size of the knowledge bases \citep{navigli2014wsd}. 

\paragraph{Knowledge-based approaches for WSD}

Knowledge-based approaches leverage part of the graph structure of word-sense inventories, e.g. WordNet, BabelNet, where words connect with all their senses. By injecting the context of a word into the graph will slightly change the graph structure, and affect the probability distribution of senses of the word in the graph, which can be computed by Personized PageRank algorithm \citep{RandomWalksWSD14}. The sense with the highest probability will be selected. 
This approach can be improved by connecting word-sense inventory with large web texts. For example, \cite{KBWSD14} applied  clique approximation algorithms, by utilizing BabelNet  \citep{Navigli21BabelNet}, a knowledge base that seamlessly integrates WordNet with Wikipedia.  

From the game theoretical perspective \citep{vonNeumann1947}, a word can be viewed as a player, and its possible senses as strategies that the player can choose, to maximize a utility function \citep{gameWSD19}. Precisely, let $W=\{w_1, \dots, w_n\}$ be the set of the content words in text $T$, $S_i=\{s_{1}, \dots, s_{m_i}\}$ be the set of senses of $w_i$, $\mathcal{S}=\bigcup S_i$  is the set of all the
strategies of the games. The strategy space of a player $w_i$ is represented as a probabilistic distribution $\mathbf{x}_i$. The way how the context determines senses of words is simulated by interactions between two words $w_i$ and $w_j$ through a utility matrix $Z$. The cell $z_{r,t}$ represents the utility value when $w_i$ chooses the $r^{th}$ strategy and $w_j$ chooses the $t^{th}$ strategy. The value of one sense's strategy is related to its partners, in the following three aspects: word similarity, word-sense similarity, and their sense distributions, and computed in the manner similar to the attention mechanism.  

\paragraph{Supervised deep learning for WSD}

Supervised deep learning approaches frame WSD as a multi-classification task -- classifying a word $w$ plus its context $C$ into one of its word-senses $s$, using an annotated corpus $\mathcal{D}$, in the form of a list of triples $<w,c,s>$, and realized by supervised deep-learning neural networks \citep{kageback16WSD,Raganato2017wsd,Mehler2018-fastsense}. 

The straightforward way of the supervised deep-learning approach is to compare the similarity between the contextualized embedding of a word in the testing context and contextualized embeddings of words in the annotated corpus, and choose the most similar one, either by feed-forward networks \citep{hadiwinoto_navigli_2019}, or transformers \citep{bevilacqua-navigli-2019-quasi}, to 
minimize a loss function $\mathcal{L}(w,c,s)$. In these approaches, word senses are treated as discrete class labels. This may cause poor performance on low frequency senses. To overcome this limitation, \cite{kumar-etal-2019-zero} explicitly computed word sense embeddings by applying embedding methods for the hypernym structure of the WordNet, then trained an attentive BiLSTM to learn the context embedding of a word to its sense embedding. \cite{scarlini_gloss2020}
computed contextualized sense embeddings by utilizing a variety of resources, such as SemCor, gloss in WordNet, SyntagNet \citep{maru_wsd19}, UKB \citep{RandomWalksWSD14}, and BERT \citep{devlin2018}. \cite{Loureiro19} computed sense embeddings by fully utilizing relations in WordNet, and achieved very competitive performance. Using explicit sense embeddings, \cite{navigli-2020-breaking} successfully reached over 80\% F1 score for WSD.  
\cite{barba2021consec} is able to choose the most important context definition for the target word. Their method inherits the idea of the game-theoretic WSD approach by using a feedback loop to consider the explicit senses of nearby words. 
Supervised neural WSD approaches outperform knowledge based approaches, in part because they are able to leverage both knowledge and annotated corpus, while knowledge-based approaches can not fully use annotated corpus.
 
\subsection{Neuosymbolic Unification}
 
Both knowledge-based and supervised deep-learning WSD approaches have two assumptions as follows: (1) word senses are 
opaque classes, (2) a sense inventory has a fixed taxonomy 
\citep{navigli21survey}. Consequently, in knowledge-based WSD approaches, word senses are represented by probabilistic distributions; in supervised WSD approaches, word senses are represented by latent vector embeddings. However, the two assumptions are somehow incompatible with the existence of a symbolic sense inventory -- if a sense inventory has a well-structured and fixed taxonomy, why senses are opaque classes in both approaches? Such incompatibility lies in the discrepancy between the continuous numeric sense representation and the discrete symbolic sense representation -- The continuous numeric representation, either as a probabilistic distribution or as a latent vector, cannot explicitly represent the well-defined symbolic taxonomy structure. This incompatibility will be resolved, if word sense embedding can precisely encode the discrete symbolic fixed taxonomy. 

By utilising geometric methods, \cite{dong19iclr} have precisely injected a large tree-structured taxonomy of sense in WordNet-3.0 into pre-trained word embeddings, resulting in a configuration of nested low-dimensional balls, which is successfully applied for the Triple classification task of large scaled knowledge graphs   \citep{dong19}. A vector sense embedding of a word is enlarged into as a ball, whose radius is geometrically computed to strictly satisfy two conditions as follows: (1) balls of sibling senses are disconnected from each other; (2) balls of child and parent senses are precisely nested -- the ball of a child sense is inside the ball of its parent sense. 
Different from other neurosymbolic representations \citep{GarcezLG2009,neuralSymbolSurvey17}, \cite{dong19iclr}'s  neurosymbolic representation unifies numerical vector embeddings and symbolic structures into one representation without loss. This enables the model to inherit good features from both neural computing and symbolic reasoning \citep{DongSpringer21}. Here, we apply the pre-trained $n$-balls for WSD, this way, senses are no more opaque classes.  

\section{{\em Dart4WSD}: A neurosymbolic Darter}

\begin{figure}[t] 
\centering
\includegraphics[width=1\linewidth]{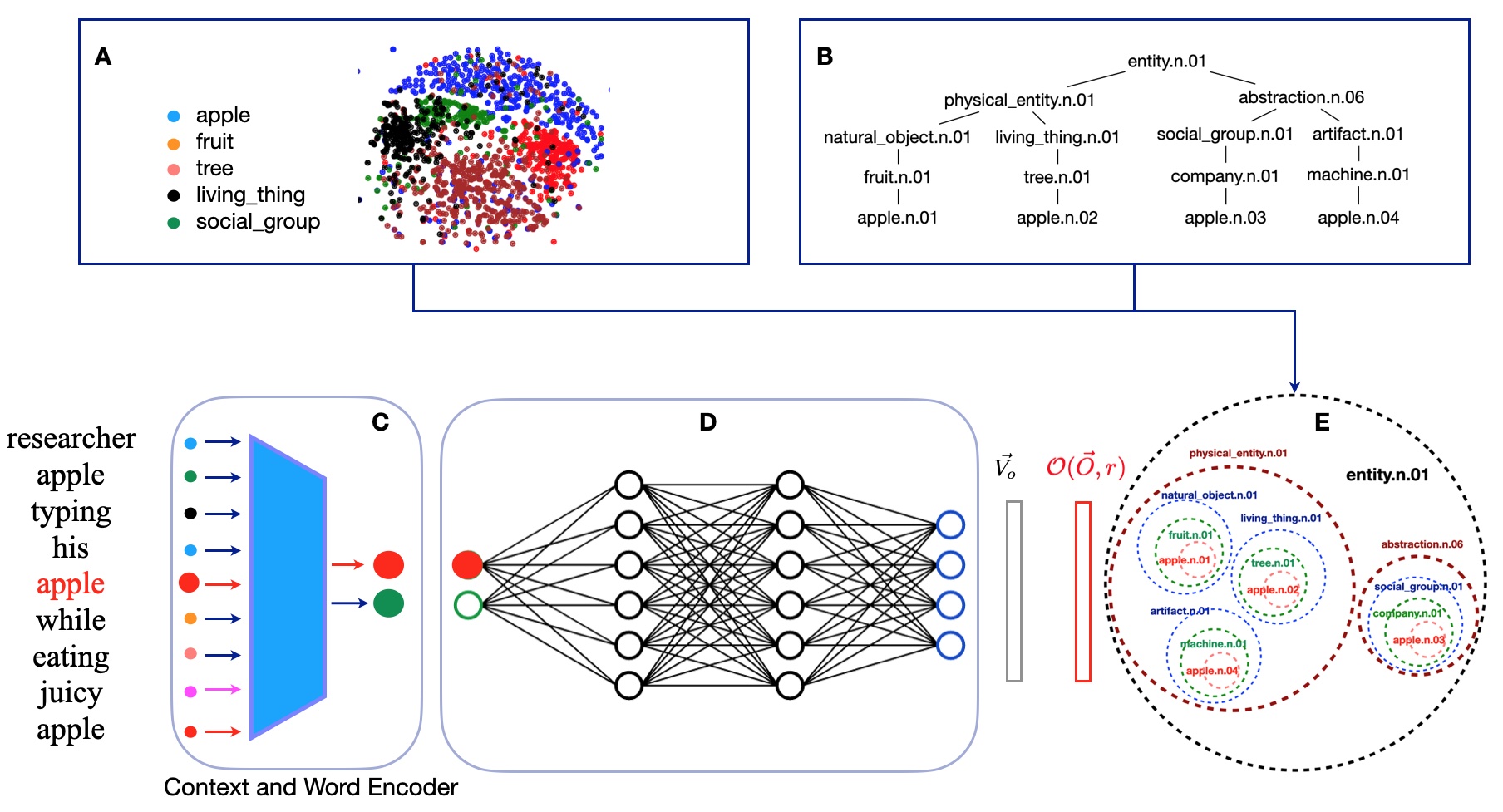}
  \caption{The supervised learning architecture of {\em Dart4WSD}: ({\bf A})  word embeddings; ({\bf B})  the fixed word-sense taxonomy extracted from a sense inventory; ({\bf C})  a neural-network that learned contextualized word embeddings; ({\bf D}) a neural network that learns to map a word in a context to its word-sense ball embedding; ({\bf E})  the neurosymbolic nested ball embeddings of word-senses.}
\label{architecture_diagram}
\end{figure}

Our {\em Dart4WSD} is a novel supervised learning methodology for Word-Sense Disambiguation, with the novelty that senses are embedded as regions in vector space and that these region embeddings explicitly represent a fixed taxonomy in a sense inventory and well-preserve pre-train vector embeddings. {\em Dart4WSD} trains a neural-network to learn the intended sense of a word in a given context, whose general architecture consists of five components; word embedding, a fixed sense inventory, a network that learns contextualized word embedding, a network that transforms the contextualized word embedding to a location in the neurosymbolic region, as illustrated in Figure~\ref{architecture_diagram}.

\subsection{Notations used in {\em Dart4WSD}}

Let $w$ and $\overrightarrow{w}$ be a word and its vector word-embedding, respectively, $C$ represent a context; $\overrightarrow{w}_C$ be a vector embedding of word $w$ in the context $C$. Let $\overrightarrow{V}$ be the output of our neural network, with the input  $\overrightarrow{w}_C$, that is, $\overrightarrow{V} =N\!N(\overrightarrow{w}_C)$. Let $w$ have $k$ different senses in the inventory $\mathcal{S}_w=\{S_1^w, \dots, S_k^w\}$, and $\mathcal{O}[S_i^w]$ be the ball embedding of $S_i^w$, with the central point $\overrightarrow{O}[S_i^w]$. {\em Dart4WSD}  learns the mapping from  $\overrightarrow{w}_C$ to $\overrightarrow{O}[S_i^w]$, the centre of $\mathcal{O}[S_i^w]$.

\subsection{The task formulation for {\em Dart4WSD}}

\begin{figure}[t] 
\centering
\includegraphics[width=0.9\linewidth]{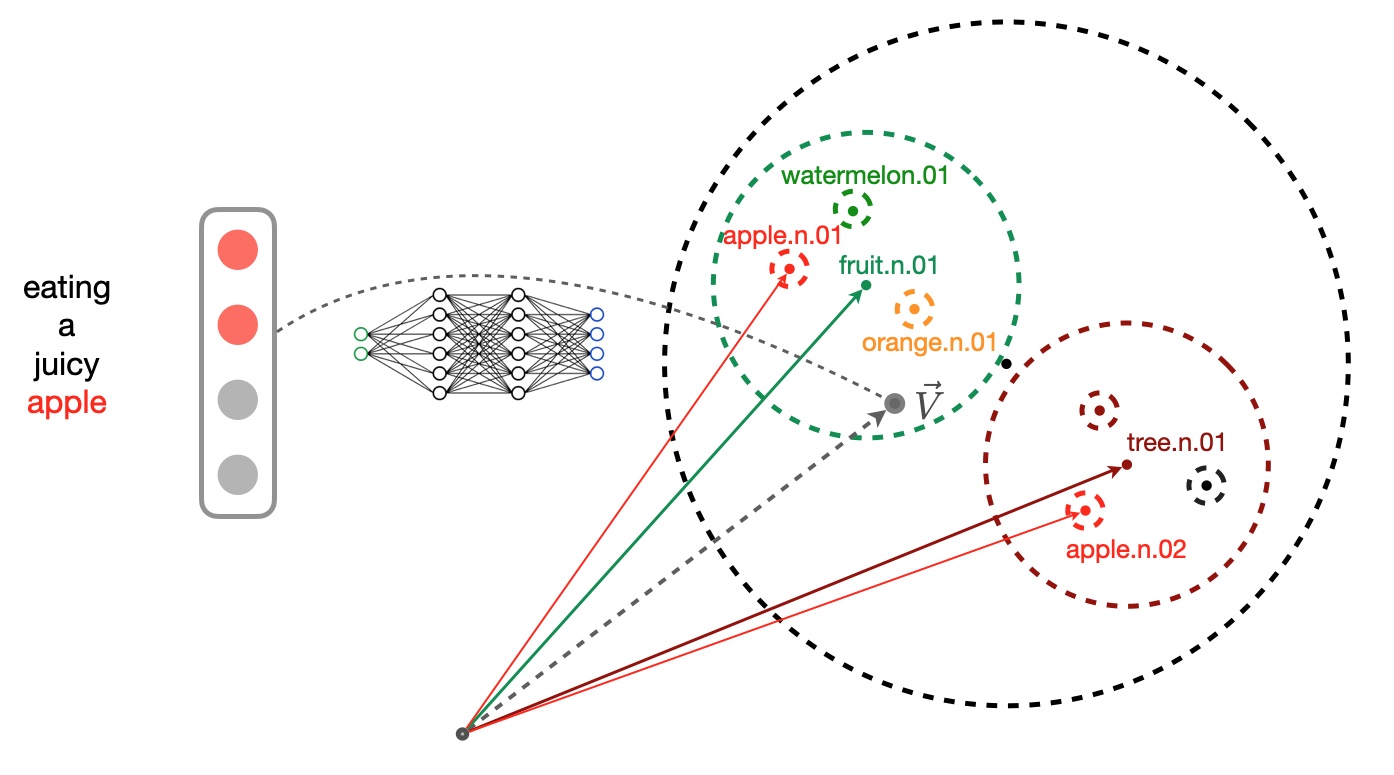}
\caption{A novel method to choose senses by carrying out reasoning with hypernym relations in the embedding space.}
\label{choose_sense}
\end{figure}

Given an annotated corpus $\mathcal{D}$, we train a neural network $N\!N$, with a loss function  $\mathcal{L}(N\!N(\overrightarrow{w}_C), \mathcal{O}[S_i^w])$ that improves the shooting technique of $N\!N$ that most of its output vectors are located inside balls of target senses. For the ease of the experiments, we use the well-known cosine similarity as an  approximation, that is, $\mathcal{L}(N\!N(\overrightarrow{w}_C), \mathcal{O}[S_i^w])\approx \cos(\overrightarrow{w}_C), \overrightarrow{O}[S_i^w]))$. This approximation works well, when balls of sibling senses in the inventory are of the similar size. For example, in Figure~\ref{choose_sense}, the apple.n.01, orange.n.01, and watermelon.n.01, three child senses of fruit.n.01, are embedded as balls with similar sizes; fruit.n.01 and tree.n.01 are siblings at the upper level in the inventory, and also embedded in the similar size. To correctly determine that the word {\em apple} in the phrase {\em eating a juicy apple}, the neural-network shall map the contextualized word embedding ($\overrightarrow{apple}_{\mbox{eating a juicy}}$) to a vector inside the ball of the sense fruit.n.01 ($\mathcal{O}[S_{1}^{fruit}]$). Then, the sense apple.n.01 inside the fruit.n.01 will be chosen as the target sense. Using upper category information for WSD in the embedding space has been proposed in \citep{WSDBLC06,vial-etal-2019-sense}, however, using regions as sense embedding distinguishes our model from these work. With regions, our model is able to reason with the symbolic hypernym relations in the embedding space. This greatly improves the performance of WSD. 

It is reasonable to argue that the context information {\em eating a juicy\dots} shall not provide information to direct the word embedding of {\em apple} exactly to the ball embedding of apple.n.01, as {\em eating a juicy orange} and {\em eating a juicy watermelon} are as meaningful as {\em eating a juicy apple}. We argue that this context information shall direct the word embedding of {\em apple} towards the sense embedding of its direct upper hypernym sense, here, fruit.n.01, and deviate from direct upper hypernym balls of its other senses, here, tree.n.01. When balls of fruit.n.01 and tree.n.01 are of similar sizes, the cosine value between the vector of network out $\overrightarrow{V}$ and the centre vector of direct upper hypernym shall be a good approximation, as illustrated in in Figure~\ref{choose_sense}. To justify this, we conducted comparative experiments (Experiment 2 vs. Experiment 4): one neural-network is trained by learning the mapping from a contextualized word ({\em apple} in {\em eating a juicy apple}) to the centre vector of its sense ball (apple.n.01); the other neural-network is trained by learning the mapping from a contextualized word to the centre vector of its direct upper hypernym sense ball (fruit.n.01). Our experiment results demonstrate no difference, in terms of the F1 score, in all six groups of evaluation data-sets. 

We formalize our novel sense selection strategy as follows:  Let $H_1(S_i^w)$ be the direct upper hypernym of $S_i^w$ in the inventory. We assume that there are no two $S_i^w$ and $S_j^w$ have the same direct upper hypernym, that is, $H_1(S_i^w)\neq H_1(S_j^w)$, if $S_i^w\neq S_j^w$. 
Then, the sense of $w$, whose $\overrightarrow{O}[H_1(S_i^w)]$ (the centre vector of the ball of the direct upper hypernym of $w$) has the largest cosine value with $\overrightarrow{V}$, will be selected as the sense of $w$ in the current context. 

$$S^w  =  \arg\max_{S_i^w\in\mathcal{S}_w}\cos(\overrightarrow{V}, \overrightarrow{O}[H_1(S_i^w)])$$ 

\begin{figure}[t] 
\centering
\includegraphics[width=1\linewidth]{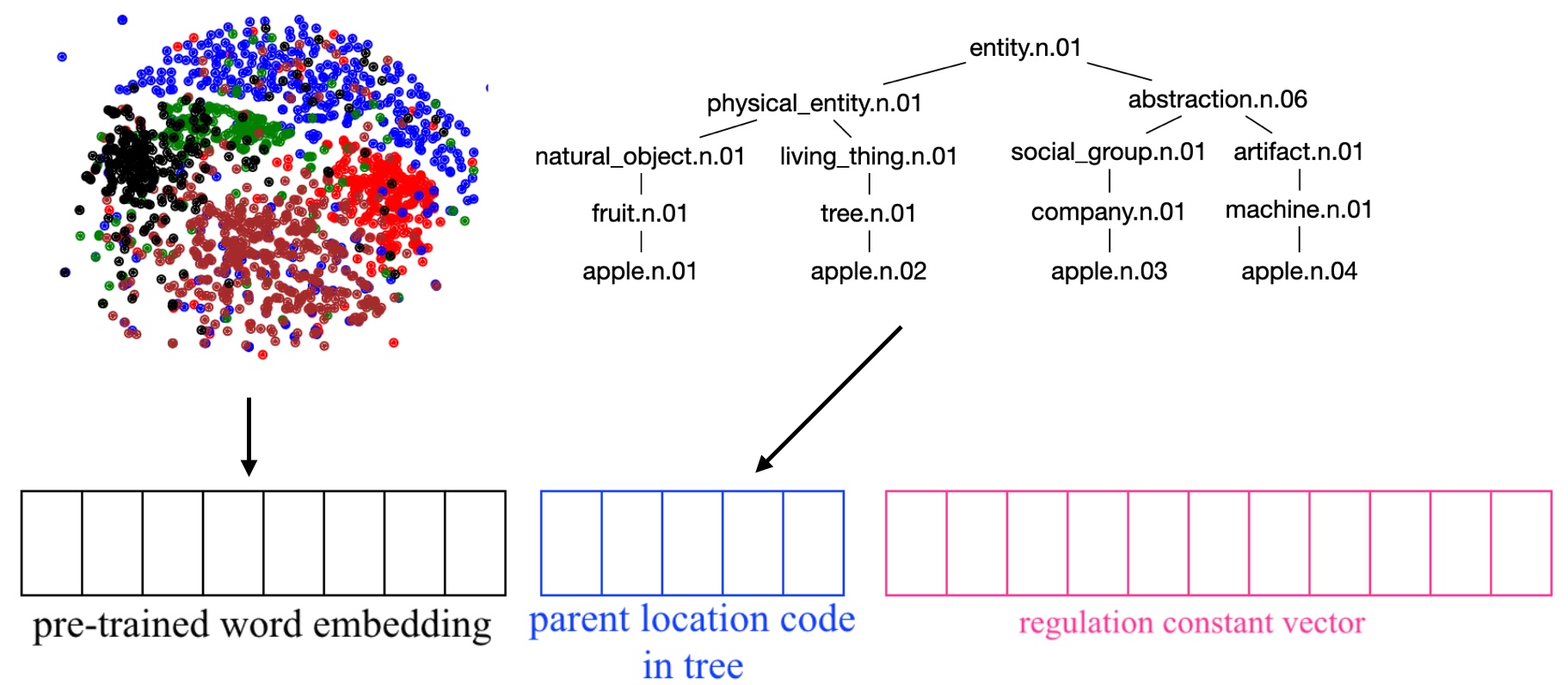}
\caption{Pre-trained word-embeddings are well-preserved, as the first part of the centre vector of an $n$-ball.}
\label{ball_center}
\end{figure}
 
\subsection{The neurosymbolic board for senses}
The centre representation of {\em Dart4WSD} is a neurosymbolic embedding of word-senses \citep{dong19}. The representation is neurosymbolic, in the sense that it unifies pre-trained word-embeddings and a symbolic tree structure of word-senses in a geometric manner, as follows: Each word-sense is represented by an $n$-dimensional ball. The pre-trained word-embeddings are well-preserved as a part of the centre vectors of balls, as illustrated in Figure~\ref{ball_center}. In this way, a sense ball also carries latent feature information learned from traditional deep-learning approaches. The symbolic tree structure is precisely represented as topological relations among balls, as illustrated in Figure~\ref{choose_sense}. Child-parent relations between senses are precisely encoded as the ball of the child sense is located inside the ball of the parent sense; sibling relations are precisely encoded as balls disconnecting from each other and being inside the same parent ball. The ball embedding of senses can be understood as an extension of vectorial embedding of senses, such that a vectorial sense embedding is a ball with the radius of zero.  

\begin{figure} 
\includegraphics[width=\linewidth]{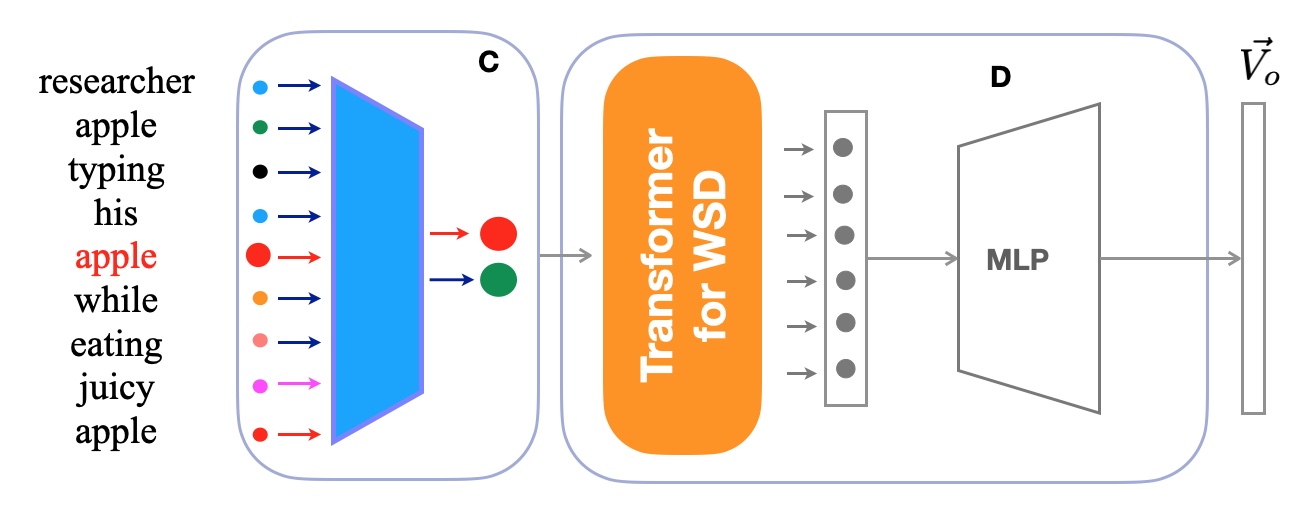}
  \caption{A transformer architecture for \textit{Dart4WSD}.}
\label{transform_dart}
\end{figure} 

\subsection{A supervised learning process for the mapping of contextualized word embeddings with sense balls}

The Transformer architecture was originally designed for sequence to sequence task \citep{vaswani2017attention}, and has been applied in a variety of fields \citep{lin2021survey}. It can be used as a universal approximator of sequence-to-sequence functions 
\citep{YunBRRK20}. We use a Transformer architecture to learn the mapping from the contextualized words to balls of their target senses, as illustrated in Figure~\ref{transform_dart}.
Given a sentence $s$, we transform it into a list of tokens $(t_{1}, t_{2}, t_{3} ... t_{m})$, then, replace each token with vector, $(\overrightarrow{t}_{1}, \overrightarrow{t}_{2}, \overrightarrow{t}_{3} ... \overrightarrow{t}_{m})$, using pre-trained word embeddings, e.g., GLOVE \citep{pennington2014glove}. 
Instead of using a network to learn contextualized word embedding, we use the average word-embedding in the $k$-window of a word as the context vector of this word (maximum $k$ words before and after a word).  Formally, the context vector the word $t_i$, written as $\overrightarrow{C}_{t_i}$, is the average of $\overrightarrow{t}_{i-k}, \dots, \overrightarrow{t}_{i-2}, \overrightarrow{t}_{i-1}, \overrightarrow{t}_{i+1},  \overrightarrow{t}_{i+2}, \dots,\overrightarrow{t}_{i+k}$ ($0<i-k$ and $i+k\le m$).  
We feed $\overrightarrow{t}_{i}$ and $\overrightarrow{C}_{t_i}$ into a Transformer, whose outputs are fed into a two-layered perceptron as follows. 
$$
    \overrightarrow{V} = Linear ( Relu (Linear(Transformer (\overrightarrow{t}_{i}, \overrightarrow{C}_{t_i}))))
$$

\begin{table}
\centering
\begin{tabular}{ |l|c|c|c|c|c| } 
\hline 
     & \#senses &\#{\bf $n$-ball}& ball ratio & \#{\bf $n$-ball L1}& L1 ratio\\\hline\hline
 {\bf SemCor} & 25845 &{\bf 15025}&$58.14\%$&{\bf 5799}&$22.44\%$ \\  
 {\bf SemCor+OMSTI} & 26265 & {\bf 15298}&$58.25\%$ &{\bf 5852}&$22.28\%$ \\\hline 
\end{tabular}
\caption{The statistics of senses in our experiments. The ball ratio is the ratio of \#$n${\-ball} to \#senses, the L1 ratio is the ratio of \#$n${\bf -ball L1} to \#senses.}
\label{stat_sense} 
\end{table}
\section{Experiments}

\subsection{Datasets}
We use the pre-trained $n$-ball embeddings  \citep{dong19iclr}. It consists of  {\bf 47,634} $n$-ball embeddings that precisely encode a large hypernym structure of word-senses in Word-Net 3.0 \citep{miller1993semantic}. One part of the central vector of the ball is the  GLOVE 50D word embedding \citep{pennington2014glove}. This restricts our experiments to be focused on English and use GLOVE 50D for the word embedding. 
\begin{table}[t]
\centering
\begin{tabular}{ |l|r|c|c|c|c| } 
\hline 
     & \#training &\#{\bf $n$-ball} &ball ratio& \#{\bf $n$-ball L1}&L1 ratio \\\hline\hline
 {\bf SemCor} & 224415 &{\bf 156483}&$69.73\%$&{\bf\underline{137552}} & $61.29\%$ \\  
 {\bf SemCor+OMSTI} & 1135547 &{\bf 837147}&$73.72\%$&{\bf 740432}&$65.20\%$\\\hline
\end{tabular}
\caption{The statistics of the numbers of training records. Here, the ball ratio is the ratio of \#$n$-ball training records to the number of senses in training records (\#training), the L1 ratio is the ratio of the number of direct hypernyms (\#L1 level) to \#training.}
\label{stat_records} 
\end{table}
\begin{table}[t] 
\begin{center}
\begin{tabular}{ |l|c|c|c|c|c|c| } 
\hline 
     & \#senses &\#test& \#$n$-ball&ball ratio&\#{\bf $n$-ball test}& test ratio\\\hline\hline
 {\bf Senseval-2} &1194& 2275 & 711&$59.55\%$&{\bf 1459}&$64.13\%$\\ {\bf Senseval-3}&1086& 1832 & 780& $71.82\%$&{\bf  1341}&$73.20\%$\\ 
  {\bf SemEval-07} & 352 &449&327&$92.90\%$&{\bf  420}&$93.54\%$\\  
  {\bf SemEval-13} &767 &1621&669&$87.22\%$&{\bf  1435} &$88.53\%$\\ 
  {\bf SemEval-15} & 524 &1013&350&$66.79\%$&{\bf 712}&$70.29\%$\\ 
  {\bf ALL}  & 3217 &7181&2251&$69.97\%$ &{\bf 5358}&$74.61\%$\\\hline 
  \end{tabular}
\caption{The statistics of the numbers of senses. The ball ratio is the ratio of \#$n$-ball to \#senses; The test ratio is the ratio of \#$n${\bf -ball test} to \#test.}
\label{evaluation_files0}
\end{center}
\end{table}
\begin{table}[h]
\begin{center}
\begin{tabular}{ |l|c|c|c|c|c| } 
\hline 
     &\#$n$-ball Lx&\#leaf&ball ratio &{\bf \#Lx test}&test ratio\\\hline\hline
 {\bf Senseval-2 L1} &478&643 &$\frac{643}{711}=90.44\%$ & 1155&$50.77\%$\\ 
  {\bf Senseval-2 L2} &316&653&$91.84\%$ &1341& $58.95\%$\\ 
  {\bf Senseval-2 L3} &224 &650&$91.42\%$&1285& $56.48\%$\\
  {\bf Senseval-2 L4}  &172 &652&$91.70\%$ &1247& $54.81\%$\\\hline 
  {\bf Senseval-3 L1}&536& 720 &  $92.30\%$& 1184&$64.63\%$\\
  {\bf Senseval-3 L2} & 380&750&$96.15\%$&1256&$68.56\%$\\
  {\bf Senseval-3 L3} &256&738 &$94.62\%$&1227&$66.98\%$\\
  {\bf Senseval-3 L4} &203&736&$94.36\%$&1198&$65.39\%$\\\hline 
  {\bf SemEval-07 L1} &245 &313&$95.72\%$& 394&$87.75\%$\\
  {\bf SemEval-07 L2} &178 &314 &$96.02\%$& 405&$90.20\%$\\
  {\bf SemEval-07 L3} &140 &313&$95.72\%$& 403&$89.75\%$\\
  {\bf SemEval-07 L4} &114 &314&$96.02\%$& 402&$89.53\%$\\\hline  
  {\bf SemEval-13 L1} &408 &571&$85.35\%$&1155 &$71.25\%$\\
  {\bf SemEval-13 L2} &261&603& $90.13\%$ &1261 &$77.79\%$\\
  {\bf SemEval-13 L3} &151&592&$88.49\%$ &1160&$71.57\%$\\
  {\bf SemEval-13 L4} &87&572&$85.50\%$ &1143&$71.51\%$\\\hline 
  {\bf SemEval-15 L1} &239 &300&$85.71\%$&606&$59.82\%$\\
  {\bf SemEval-15 L2} &197&311&$88.86\%$&640& $63.18\%$\\
  {\bf SemEval-15 L3} &161 &341&$97.43\%$&671&$66.24\%$\\
  {\bf SemEval-15 L4} &125&316&$90.29\%$&618&$61.01\%$ \\\hline  
  {\bf ALL L1}&1270 &2056&$91.34\%$&4485& $62.46\%$ \\
  {\bf ALL L2} &760 &2108&$93.65\%$&4895&$68.17\%$\\
  {\bf ALL L3}&480 &2113&$93.87\%$&4739& $65.99\%$\\
  {\bf ALL L4}&330 &2093&$92.98\%$&4601&$64.01\%$\\\hline
\end{tabular}
\caption{The statistics of the numbers of senses. The ball ratio is the ratio of \#leaf to \#n-ball in Table~\ref{evaluation_files0}; the test ratio is the ratio of \#{\bf Lx test} to \#test in Table~\ref{evaluation_files0}. The first row is read as {\em the set of direct hypernyms in {\bf Senseval-2 L1} data-set has 478 senses, covering 643 leaf senses. These 643 senses cover 90.44\% senses in the {\bf Senseval-2} data-set; these 478 senses have 1155 testing records, that is 55.77\% of the testing records in in the {\bf Senseval-2} data-set.}}
\label{evaluation_files}
\end{center}
\end{table}
Based on two benchmark WSD corpus SemCor and SemCor$+$OMSTI, we create four training data-sets, as follows: (1) {\bf SemCor-$n$ball}, (2) {\bf SemCor$+$OMSTI-$n$ball}, (3) {\bf SemCor-$n$ball-L1}, and (4) {\bf SemCor$+$OMSTI-$n$ball-L1}. 
The four data-sets are created in the following way: Firstly, we transform training data into the form as follows. $$(\mbox{sense}, \mbox{a list of word}, \mbox{the index that points to the word(s) of the sense})$$
For example, 
$$(\mbox{`aim.n.02'}, \mbox{[`have', `you', `set', `specific', `objectives', `for', `your', `career']}, \mbox{[4]})$$
which means that the word pointed by the index 4, that is the word `objectives', should have the sense `aim.n.02'. 
The first two data-sets {\bf SemCor-$n$ball} and {\bf SemCor$+$OMSTI-$n$ball} are extracted from SemCor and SemCor$+$OMSTI with the criteria that senses have $n$-ball embeddings. That is, if `aim.n.02' has an $n$-ball embedding, this piece of training record will be selected.  The other two data-sets are created, by setting each target sense in the first two datasets with its direct hypernym. If this hypernym has $n$-ball embedding, the training record will be selected.  
For example, `aim.n.02' has an hypernym path in WordNet-3.0, as follows: [`aim.n.02', `goal.n.01', `content.n.05', `cognition.n.01', \dots]. Its direct hypernym is `goal.n.01'. If it has an $n$-ball embedding, the following training record will be added into the corresponding {\bf -L1} data-set. 
$$(\mbox{`goal.n.01'}, \mbox{`aim.n.02'},\mbox{[`have', `you', `set', `specific', `objectives', `for', \dots]}, \mbox{[4]})$$
Table~\ref{stat_sense} and Table~\ref{stat_records} list the statistics of senses and training records. The SemCor data-set has 25845 senses (\#senses) and  224415 training records, among which 15025 senses exist in the pre-trained $n$-ball embedding (\#$n$-ball). Senses in 156483 training records have $n$-ball embeddings. 57799 of the direct hypernym senses have $n$-ball embedding (\#$n$-ball L1), and correponding to 137552 training records. 
This way, our experiments use around $70\%$ training data in the SemCor and the SemCor$+$OMSTI data-sets, covering around $58\%$ senses of them. 

For testing, we create $6\times 5 =30$ data-sets from all standard data-sets, namely, {\bf Senseval-2}, {\bf Senseval-3}, {\bf SemEval-07}, {\bf SemEval-13}, {\bf SemEval-15}, {\bf ALL} \citep{raganato2017word}. From each data-set {\bf E}$\in$\{{\bf Senseval-2}, {\bf Senseval-3}, {\bf SemEval-07}, {\bf SemEval-13}, {\bf SemEval-15}, {\bf ALL}\}, we  derive 5 testing data-sets as follows: {\bf E-$n$ball}, {\bf E-$n$ball-L1}, \dots, {\bf E-$n$ball-L4}. {\bf E-$n$ball} and {\bf E-$n$ball-L1} are created in the same way as we create training data. {\bf E-$n$ball-L$i$} ($i=2,3,4$) is created  by setting the target sense with its $i^{th}$ upper hypernym. If this hypernym has $n$-ball embedding, this training record will selected. For example, `content.n.05' is the second level hypernym of `aim.n.02' and has $n$-ball embedding, so, the following testing record is added into {\bf E-$n$ball-L2}.
$$(\mbox{`content.n.05'}, \mbox{`aim.n.02'},\mbox{[`have', `you', `set', `specific', `objectives', `for', \dots]}, \mbox{[4]})$$
For evaluation, we use the F1 calculation software in the standard WSD corpus, downloaded from \url{http://lcl.uniroma1.it/wsdeval/home}. 
\subsection{The aims of  experiments}

We conducted a series of experiments to answer  questions as follows.
\begin{enumerate}
    \item How good is our {\em Dart4WSD} in the task of mapping contextualized word {\bf vector} to sense {\bf vectors}? 
    \item How is the performance of {\em Dart4WSD}, if measured by whether it hits the direct upper hypernym of the target sense (here, $n$-dimensional balls)?
    \item  Would performances be improved, if we enlarge sizes of target balls? 
    \item  Will the performance be improved in the testing phase, if  in the learning phase it maps to $n$-balls of direct upper hypernym senses? 
    \item  Is {\em Dart4WSD} data-hungry?
\end{enumerate} 
\begin{table}[t]
\begin{center}
\begin{tabular}{ |c|c|c|c|c|c| } 
\hline 
      {\bf Senseval-2}&{\bf Senseval-3} &{\bf SemEval-07}&{\bf SemEval-13}&  {\bf SemEval-15}&  {\bf ALL} \\\hline
    $37.3\%$ & $36.4\%$ & $36.7\%$ &$35.5\%$ &$35.5\%$ &  $36.3\%$\\ 
      \hline 
  \end{tabular}
\caption{The performance of hitting the sense vectors by using the model trained by {\bf SemCor-$n$ball} data-set.}
\label{result_spoint}
\end{center}
\end{table}
\subsection{Experiments and Results}
\subsubsection{Experiment 1} To answer the first question, we used the {\bf SemCor-$n$ball} data-set to train our {\em Dart4WSD} neural-network. It learns to map from contextualized word embeddings to center vectors of sense $n$-balls. The performances range from F1 score 35.5\% to 37.3\% in all the testing data, as illustrated in the column {\bf sense vector} of Table~\ref{result_spoint}. Compared with the current best result $80\%$ \citep{navigli-2020-breaking}, our performance is poor. 
\subsubsection{Experiment 2} To answer the second question, we used the trained model in {\rm Experiment 1}, and evaluate whether it successfully hit the ball of the direct upper hypernym senses. The F1 scores range from $90\%$ to $100\%$ in all the testing data, as illustrated in the column {\bf L1} of Table~\ref{tbl:result}. This performance greatly outperform  the current best result, and break the performance ceiling (a bit above 90\%) of traditional deep-learning approaches \citep{raganato2017word}.
\subsubsection{Experiment 3} For the third question, we used the trained model in {\rm Experiment 1}, and evaluate this performances in hitting balls of the second to the fourth upper hypernym senses. There are very little improvement in terms of the F1 scores, as illustrated in the columns from {\rm L2} to {\rm L4} of Table~\ref{tbl:result}.  Ball sizes of upper level hypernyms are larger than ball sizes of lower level hypernyms, thus, they are easier to hit. This will increase the value of recall. On the other hand, increased ball sizes may include unintended senses. This will reduce the value of precision. In our experiment results, the two are well balanced.   
\begin{table}[t] 
  \centering
  \begin{tabular}{|c|c|c|c|c|c|c|c|c|c|}
    \hline
    \multicolumn{2}{|c|}{} &\multicolumn{4}{|c|}{\bf Senseval-2}  &\multicolumn{4}{|c|}{\bf Senseval-3} \\
     \cline{3-10}
     \multicolumn{2}{|c|}{}  &  {\bf L1} & {\rm L2} & {\rm L3} & {\rm L4}&{\bf L1} & {\rm L2} & {\rm L3} & {\rm L4}\\\hline 
     \multicolumn{2}{|c|}{{\bf SemCor}}&  {\bf 94.4\%}  & $95.6\%$  &$95.2\%$& $95.0\%$  &  {\bf 92.4\%}  & $92.8\%$  &$92.3\%$& $92.1\%$ \\\cline{3-10}
   \multicolumn{2}{|c|}{{\bf \underline{SemCor L1}}}&  {\bf 94.4\%}  & $95.6\%$  &$95.2\%$& $95.0\%$   &  {\bf 92.4\%}  & $92.8\%$  &$92.3\%$& $92.1\%$\\\cline{3-10}
    \multicolumn{2}{|c|}{{\bf SemCor+OMSTI}}& $94.4\%$  & $95.6\%$  &$95.2\%$& $95.6\%$  &  $92.4\%$  & $92.8\%$  &$92.3\%$& $92.1\%$\\\cline{3-10}
   \multicolumn{2}{|c|}{{\bf SemCor+OMSTI L1}}&$94.4\%$  & $95.6\%$  &$95.2\%$& $95.0\%$  & $92.4\%$  & $92.8\%$  &$92.3\%$& $92.1\%$\\\cline{3-10} 
   \hline  
       \multicolumn{2}{|c|}{} &\multicolumn{4}{|c|}{\bf SemEval-07}  &\multicolumn{4}{|c|}{\bf SemEval-13} \\
     \cline{3-10}
     \multicolumn{2}{|c|}{}  & {\bf L1} & {\rm L2} & {\rm L3} & {\rm L4} & {\bf L1} & {\rm L2} & {\rm L3} & {\rm L4}\\\hline  
     \multicolumn{2}{|c|}{{\bf SemCor}}& {\bf 90.1\%} &  $90.9\%$  & $91.3\%$& $91.3\%$ &  {\bf 100.0\%} &$100.0\%$  &$100.0\%$   &$100.0\%$  \\\cline{3-10}
   \multicolumn{2}{|c|}{{\bf \underline{SemCor L1}}}&   {\bf 90.1\%} &  $90.9\%$  & $91.3\%$& $91.3\%$&{\bf 100.0\%}  &$100.0\%$   &$100.0\%$ &$100.0\%$  \\\cline{3-10}
    \multicolumn{2}{|c|}{{\bf SemCor+OMSTI}} & $90.1\%$ &  $90.9\%$  & $91.3\%$& $91.3\%$ &$100.0\%$  &$100.0\%$   &$100.0\%$ &$100.0\%$  \\\cline{3-10}
   \multicolumn{2}{|c|}{{\bf SemCor+OMSTI L1}}&  $90.6\%$ &  $90.9\%$  & $91.3\%$& $91.3\%$  &$100.0\%$  &$100.0\%$   &$100.0\%$ &$100.0\%$  \\\cline{3-10} 
   \hline  
       \multicolumn{2}{|c|}{} &\multicolumn{4}{|c|}{\bf SemEval-15}&\multicolumn{4}{|c|}{\bf ALL} \\
     \cline{3-10}
     \multicolumn{2}{|c|}{}  &  {\bf L1} & L2 & {\rm L3} & {\rm L4} &  {\bf L1} & {\rm L2} & {\rm L3} & {\rm L4}\\\hline 
     \multicolumn{2}{|c|}{{\bf SemCor}}&   {\bf 90.3\%}  & $91.3\%$& $91.7\%$&  $90.9\%$ &{\bf 94.4\%}&  $95.0\%$  & $94.8\%$& $94.6\%$  \\\cline{3-10}
   \multicolumn{2}{|c|}{{\bf \underline{SemCor L1}}} &  {\bf 90.3\%}  & $91.3\%$& $91.7\%$&  $90.9\%$ &{\bf 94.4\%}&  $95.0\%$  & $94.8\%$& $94.6\%$  \\\cline{3-10}
    \multicolumn{2}{|c|}{{\bf SemCor+OMSTI}} &  $90.3\%$  & $91.3\%$& $91.7\%$&  $90.9\%$ &$94.4\%$&  $95.0\%$  & $94.8\%$& $94.6\%$  \\\cline{3-10}
   \multicolumn{2}{|c|}{{\bf SemCor+OMSTI L1}}&  $90.3\%$  & $91.3\%$& $91.7\%$&  $90.9\%$ &$94.4\%$&  $95.0\%$  & $94.8\%$& $94.6\%$  \\\cline{3-10}
   \hline  
  \end{tabular}
  \caption{F1 scores of 24 testing datasets. The F1 computed by the standard tool for WSD, which is available in the data-set download from  \url{http://lcl.uniroma1.it/wsdeval/home} .}
  \label{tbl:result}
\end{table}

\subsubsection{Experiment 4} To answer the fourth question, we used the {\bf SemCor-$n$ball-L1} data-set, and the neural-network learns to map from contextualized word embeddings to centre vectors of $n$-balls of the direct upper hypernym senses. 
There are no improvement in terms of the F1 scores, as illustrated in the rows {\bf SemCor L1} of Table~\ref{tbl:result}. This supports our assumption in Section 3.2.  

\subsubsection{Experiment 5} To answer the fifth question, we used  {\bf SemCor$+$OMSTI-$n$ball} and {\bf SemCor$+$OMSTI-$n$ball L1} data-sets, and repeated the above experiments and evaluations. There are no changes as using {\bf SemCor-$n$ball} and {\bf SemCorn-$n$ball L1} data-sets, as illustrated in Table~\ref{tbl:result}. The ceiling of traditional supervised learning for WSD is a little above $90.0\%$ in F1 score \citep{raganato2017word}, so achieving F1 score far above that level shall be ascribed to the hypernym structures that is imposed onto the embedding space. That is, {\em Dart4WSD} is no more data hungry, after being fed with sufficient precise structural knowledge, in terms of a configuration of balls, which can be understood as an Euler diagram that supports for logical deduction, for example, greek.n.01 is human.n.01, and human.n.01 is mammal.n.01, so, the ball of greek.n.01 is inside the ball of human.n.01 that is inside the ball of mammal.n.01. To deduce that greek.n.01 is mammal.n.01, we can inspect whether the ball greek.n.01 is inside the ball mammal.n.01. This way, a sense is selected, if it is inside the ball of its direct hypernym sense. This kinds of simple deduction relies on the boundary relations between balls, and not relevant to their centre points (where latent features are encoded). And this is possible, only when we precisely impose a symbolic taxonomy onto the embedding space. This boundary information greatly improves the performance of reasoning, when training data is not sufficient.  

\subsection{Summary}
These experiment results affirm the soundness of our {\em Dart4WSD} method, and suggests that the best training corpus is to use direct upper hypernym. Here, it will be the {\bf SemCor-$n$ball-L1}, because the set of direct upper hypernyms contain less number of senses than the set of their leaf senses, and achieves the same performance as by using {\bf SemCor-$n$ball}, and the best test corpus are {\bf E-$n$ball-L1}, where {\bf E}$\in$\{{\bf Senseval-2}, {\bf Senseval-3}, {\bf SemEval-07}, {\bf SemEval-13}, {\bf SemEval-15}, {\bf ALL}\}.  

\subsection{Discussions and lessons learned}

This work creates a number of issues that worth discussion and continued exploration. We used the cosine similarity in the loss function, which is easily for implementation, and works well in the current setting of experiments. However, in general, the cosine similarity cannot take the boundary information of balls into the consideration. In a full-fledged neurosymbolic approach for WSD, we shall replace the cosine similarity with spatial functions that are able to explicitly describe topological relations among regions, such as, being part of, disconnecting from, being partial overlapped with. 

One assumption of our approach is that senses of word shall have different direct upper hypernyms, so, we can use balls of direct upper hypernyms. This assumption holds for nouns in most of the cases, but, might not hold for verbs. For example, fly.v.01 ({\em travel through the air; be airborne}) and fly.v.06 ({\em be dispersed or disseminated}) are both senses the word fly, they share the same direct upper hypernym travel.n.01 ({\em change location; move, travel, or proceed, also metaphorically}). In this case, using direct upper hypernym is not sufficient to disambiguate between fly.v.01 and fly.v.06. We may need to update the existing $n$-ball embedding, to meet the need of WSD, for example, to enlarge sizes of balls of leaf nodes in the sense inventory, while keeping the topological structure. 

The Transformer using the average neighbourhood word embeddings as the context demonstrated unsatisfactory performances in the experiments. On one hand, this supports the power of our novel neurosymbolic method -- it achieves over $90\%$ F1 score when its output is mapped with balls of senses. On the other hand, this creates room for improvements. For example, it might be promising to promote current vector-based WSD systems into ball-based, in which a neural-network is used to compute a contextualized word embedding, and another neural link is  established between contextualized word embeddings and vectorial sense embeddings. We may create a taxonomy of senses and impose it onto these vectorial sense embeddings, and promote them into $n$-ball embeddings. This  will develop more powerful neurosymbolic darting for WSD.

\section{Conclusions, applications, on-going work}

We presented a novel method for Word-Sense Disambiguation (WSD), which utilizes neurosymbolic representation of word-senses. In this representation, word-senses are represented balls in the vector space, such that the inclusion relations precisely encode a symbolic hypernym structure of word-senses, and the pre-trained word-embeddings are well-preserved in the centre of balls. A supervised deep-learning process is carried out by learning the mapping between contextualized word-embeddings and balls of senses. This process can be vividly metaphored as the game of darting. A neural-network shoots the arrow of a word in a context to the region of its sense.  This is much easier than shooting to a point, as tried in most of other supervised approaches. Experiments show that our novel method demands neither huge amount of training data, nor high techniques of shooting.

Our current experiments are limited by the pre-trained $n$-ball embeddings, which covers around $58\%$ senses and $70\%$ sentences in the training data, and  $59\%$-$72\%$ senses and $64\%$-$93\%$ sentences in the testing data. We will extend $n$-ball embeddings to cover all these missing senses and related hypernym relations. Our novel method is independent of languages, and can be applied for the WSD of languages with insufficient training corpus. In despite of these, the current system can be applied together with existing WSD systems, when words are not available in the $n$-ball embeddings.

\bibliographystyle{apalike}
\bibliography{references}
\end{document}


\title{Supplemental Material Statement for the Paper ``Word Sense Disambiguation as a Game of Neurosymbolic Darts''}
%
%
\author{First Author\inst{1}\orcidID{0000-1111-2222-3333} \and
Second Author\inst{2,3}\orcidID{1111-2222-3333-4444} \and
Third Author\inst{3}\orcidID{2222--3333-4444-5555}}
\authorrunning{F. Author et al.}
%
%
\maketitle              
%

%
%
%

Source code and the step-by-step guide to download corpus and to run coedes are available at \url{https://anonymous.4open.science/r/dart4wsd-4A0E/README.md}

Pre-trained corpus, testing corpus, and four pre-trained models are available at
\url{https://figshare.com/s/9b25aaf51299a1814a20}

\bibliographystyle{apalike}
\bibliography{references}